\documentclass[runningheads]{llncs}

\usepackage{microtype}
\usepackage{graphicx}
\usepackage{booktabs} 
\usepackage[usenames, dvipsnames, table]{xcolor}

\usepackage{mathtools}
\usepackage{algorithm}
\usepackage{algpseudocode}
\usepackage{tikz}
\usetikzlibrary{arrows,automata,positioning}
\usepackage{multirow} 
\usepackage{xspace}
\usepackage{hyperref}
\usepackage[capitalise]{cleveref}
\crefformat{equation}{(#2#1#3)}

\usepackage{comment}

\usepackage{xcolor}

\DeclarePairedDelimiterX{\norm}[1]{\lVert}{\rVert}{#1}

\newcommand{\ie}[0]{\emph{i.e.},~}
\newcommand{\eg}[0]{\emph{e.g.},~}

\usepackage{dsfont}
\usepackage{tabulary}
\usepackage{tabularx}
\usepackage{makecell}
\usepackage{wrapfig}
\usepackage[numbers,sort&compress]{natbib}

\newcommand{\ftsize}{\scriptsize}
\newcommand{\fsize}{scriptsize}
\usepackage[font=\fsize,labelfont=bf]{caption}
\usepackage[font=\fsize]{subfig}
\usepackage[skip=5pt]{caption}
\usepackage[export]{adjustbox}
\usepackage{wrapfig}

\usepackage{ragged2e}

\definecolor{orange}{rgb}{1,0.5,0}
\definecolor{lightsalmonpink}{rgb}{1.0, 0.6, 0.6}
\definecolor{verylightsalmonpink}{rgb}{0.966, 0.805, 0.797}
\definecolor{lightblue}{rgb}{0.862, 0.906, 0.984}
\definecolor{lightyellow}{rgb}{1.0, 0.945, 0.797}
\definecolor{lightgreen}{rgb}{0.835, 0.91, 0.828}
\definecolor{lightpurple}{rgb}{0.879, 0.832, 0.902}

\newcommand{\ttbf}[1]{\textbf{\texttt{#1}}}

\newcommand{\mat}[1]{\mathbf{#1}}
\newcommand{\vect}[1]{\mathbf{#1}}

\newcommand{\kw}[1]{\textsc{\MakeLowercase{#1}}}

\newcommand{\app}{\kw{CYBORGS}}

\usepackage{float}
\usepackage[accsupp]{axessibility}

\belowdisplayskip \abovedisplayskip

\newlength{\abstractReduceTop}
\newlength{\abstractReduceBot}

\newlength{\sectionReduceTop}
\newlength{\sectionReduceBot}

\newlength{\subsectionReduceTop}
\newlength{\subsectionReduceBot}

\newlength{\subsubsectionReduceTop}
\newlength{\subsubsectionReduceBot}

\newlength{\captionReduceTop}
\newlength{\captionReduceBot}

\newlength{\eqnReduceTop}
\newlength{\eqnReduceBot}

\newlength{\horSkip}
\newlength{\verSkip}

\newlength{\figureHeight}
\begin{document}
\pagestyle{headings}
\mainmatter
\def\ECCVSubNumber{5373}  

\title{CYBORGS: Contrastively Bootstrapping Object Representations by Grounding in Segmentation} 

\titlerunning{CYBORGS}
%
\author{Renhao Wang\inst{1} \and
Hang Zhao\inst{1,2,\star} \and
Yang Gao\inst{1,2,}\thanks{equal advising}}
\authorrunning{R. Wang et al.}
%
\institute{Tsinghua University
\and
Shanghai Qi Zhi Institute \\ \email{\{renwang435, zhaohang0124\}@gmail.com, 	gaoyangiiis@tsinghua.edu.cn}}
\maketitle

\begin{abstract}
Many recent approaches in contrastive learning have worked to close the gap between pretraining on iconic images like ImageNet and pretraining on complex scenes like COCO. This gap exists largely because commonly used random crop augmentations obtain semantically inconsistent content in crowded scene images of diverse objects. In this work, we propose a framework which tackles this problem via joint learning of representations and segmentation. We leverage segmentation masks to train a model with a mask-dependent contrastive loss, and use the partially trained model to bootstrap better masks. By iterating between these two components, we ground the contrastive updates in segmentation information, and simultaneously improve segmentation throughout pretraining. Experiments show our representations transfer robustly to downstream tasks in classification, detection and segmentation.\footnote{Code and pretrained models available at \url{https://github.com/renwang435/CYBORGS}}
\end{abstract}

\section{Introduction}

Many self-supervised contrastive methods have come to rival and even surpass the performance of fully supervised methods on a number of tasks, including object detection \cite{bar2021detreg, xiao2021region}, semantic segmentation \cite{hamilton2021unsupervised, Van_Gansbeke_2021_ICCV}, video understanding \cite{jabri2020space, kuang2021video}, and image classification \cite{doersch2015unsupervised, chen2020simple}. A large portion of these methods rely on random cropping to select positive pairs of image subregions for a self-supervised instance-level discrimination task. Recently, many works have found this random cropping strategy succeeds for iconic image pretraining, but struggles when applied to pretraining on complex scene images. Treating two random crops from the same image as containing semantically similar information works well for images with singular, dominant subjects, like those in ImageNet. But such an assumption inevitably fails due to inconsistent learning signals in scene images full of diverse objects~\cite{selvaraju2021casting, chen2021intriguing, liu2020self}. To address this issue, prior works have generated random crops in an object-aware manner \cite{xiao2021region, bar2021detreg, mo2021object}. By localizing objects with unsupervised algorithms (e.g. selective search), these works are able to ground random crops around singular objects, validating the assumption that such crops contain similar information about objects.

We argue that utilizing \emph{pixel-level} object information can be even more effective than detection-level boxes. By parsing random crop contents with segmentation masks, we can turn a pair of crops into a diverse wealth of similar and dissimilar object regions, facilitating contrastive self-supervised learning. To fully realize this idea in SSL frameworks, we also need to meet two important criteria. Firstly, these masks should be obtained in an \emph{unsupervised} manner. Secondly, we want to \emph{avoid preprocessing pipelines} to obtain pseudo-segmentations (\eg graph cut algorithms), which often lack awareness of object-level semantics and require human domain knowledge for good performance~\cite{henaff2021efficient, zhang2021looking}.

\begin{figure}[tb]
    \centering
    \vspace{-15pt}
    \scalebox{0.5}{
        \includegraphics[width=\linewidth]{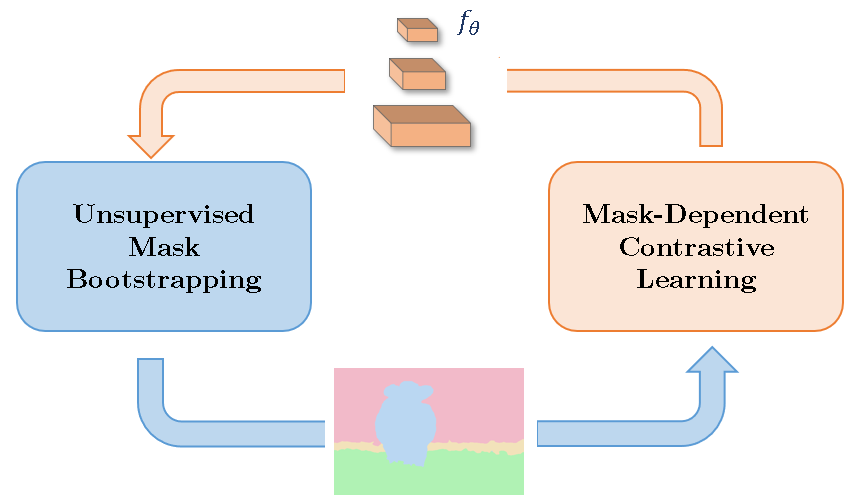}
    }
    \caption{\textbf{Mutually improving representation learning and semantic segmentation.} In the first stage, we use available segmentation masks to ground contrastive learning. In the second stage, we use representations from the backbone $f_\theta$ to bootstrap improved segmentation masks.}
    \label{fig:intro}
    \vspace{-15pt}
\end{figure}

To this end, we propose in this work to perform segmentation and concept learning \emph{jointly} (\cref{fig:intro}). In the first stage of our framework, we ground self-supervised learning with segmentation information to train a representation backbone. In a periodic second stage, we leverage these representations to bootstrap segmentation masks, which can subsequently be fed back to the first stage to further improve representations. By iterating between these two core stages, we develop representations which strongly generalize to many downstream tasks, and are especially well-aligned with object detection and segmentation. 
Furthermore, to ameliorate issues of representation collapse, we also optimize a clustering consistency objective during the first stage. We show that the formulation of this loss fits naturally within any contrastive framework, and helps improve masks more reliably between bootstrap cycles. Thus, in \textbf{C}ontrastivel\textbf{Y} \textbf{B}ootstrapping \textbf{O}bject \textbf{R}epresentations by \textbf{G}rounding in \textbf{S}egmentation (\app), our contributions are fourfold:

\begin{enumerate}
    \item
    We develop the first framework which performs end-to-end, joint self-supervised learning of object-level representations and semantic segmentation, while removing entirely the need for heuristic preprocessing of pseudo-segmentations.
    
    \item
    We show how to bootstrap segmentation masks robustly by directly clustering on feature maps obtained from a partially pretrained backbone.
    
    \item
    We demonstrate how to regularize contrastive updates in our framework with an intra-/inter-view cluster consistency loss that is well-aligned with the hyperspherically-distributed contrastive embeddings.
    
    \item
    With pretraining on complex scene images such as COCO, we demonstrate that grounding in segmentation leads to representations which transfer competitively to a diversity of downstream tasks and real-world, long-tail objects and scene semantics.

\end{enumerate}

\section{Related Work}

\looseness=-1 \paragraph{Self-supervised representation learning.} SSL methods utilize internal structure as a source of supervision to learn general representations, including auxiliary tasks such as context prediction \cite{doersch2015unsupervised}, solving jigsaws \cite{noroozi2016unsupervised}, inpainting \cite{pathak2016context}, colorization \cite{zhang2016colorful}, or orientation prediction \cite{komodakis2018unsupervised}. Most relevant to our work is contrastive learning, where the goal is to perform instance discrimination, concentrating positive pairs and separating negative pairs of feature embeddings in a latent space \cite{he2020momentum, chen2020simple, oord2018representation, grill2020bootstrap}. Despite their convincing performance on downstream tasks, the majority of current contrastive-based methods are pretrained on ImageNet, and subject to strong object-centric bias and poor visual grounding \cite{selvaraju2021casting, mo2021object, herranz2016scene, purushwalkam2020demystifying}.

\looseness=-1 To this end, a number of emerging methods examine self-supervised representaton learning on in-the-wild, scene image datasets such as COCO \cite{selvaraju2021casting, liu2020self, xie2021unsupervised}. CAST improves visual grounding by ensuring crops overlap readily with object regions identified by saliency masks, and guides representation learning using a Grad-CAM loss \cite{selvaraju2017grad, selvaraju2021casting}. ORL uses a pretrained self-supervised model to approximate object-level semantic correspondence, thus improving positive-negative identification for contrastive refinement of the pretrained model \cite{xie2021unsupervised}.

\looseness=-1 Going a step further, \app{} and other works obtain object-level semantics through pixel level pseudo-labeling \cite{henaff2021efficient, Van_Gansbeke_2021_ICCV, zhang2020hierarchical, bai2022point, hamilton2021unsupervised}. For example, DetCon \cite{henaff2021efficient} involves unsupervised preprocessing of images to obtain masks, and uses these masks to aggregate features over object regions for contrastive learning. Crucially, all other previous methods suffer from the disadvantage that mask proposals are generated i) via graph-based algorithms requiring heuristic hyperparameter decisions, and ii) only once before training, with no further learning. In contrast, by integrating object mask proposals and contrastive pretraining into the same loop, \app{} iteratively refines and improves both segmentation quality and learned representation quality, jointly.

\looseness=-1 \paragraph{Unsupervised segmentation and clustering.} The use of clustering-based approaches in SSL has a long history \cite{caron2018deep, caron2020swav, PCL}. DeepCluster is a seminal work which proposed to train a CNN by alternating between feature clustering to obtain class pseudo-labels, and learning to predict those very labels \cite{caron2018deep}. PCL and SwAV combine a clustering objective with a contrastive objective, directly encoding semantic structure learned by clustering into a latent representation space \cite{caron2020swav, PCL}. Instead of directly improving features by learning to cluster feature prototypes, \app{} primarily uses clustering as a mechanism to improve segmentation.

\looseness=-1 Indeed, clustering-based algorithms have recently found application in a number of unsupervised and self-supervised image segmentation works~\cite{hwang2019segsort, zhang2021looking, zhang2020hierarchical, cho2021picie}. Both pixel and region-level contrastive learning methods have been employed to i) improve semantic segmentation for better representation learning \cite{zhang2020hierarchical, Xiong_2021_ICCV}, and ii) vice versa \cite{hwang2019segsort, ke2021spml, cho2021picie}. To the best of our knowledge, \app{} is the first work to consider these two well-studied tasks as complementary, iteratively synergizing them together via a bootstrapping paradigm. Additionally, \app{} does not directly optimize for segmentation quality via pixel-level losses, and aims to improve segmentation strictly insofar as it aids in representation learning.

\section{CYBORGS}

We now describe the details in our proposed framework. In \cref{sec:overall}, we provide an overview of the abstractions in our work. At its core, we require iteration between two components: a contrastive objective capable of leveraging masks to train an encoder, and an unsupervised method to generate masks from a (partially) trained encoder. In \cref{sec:mask_obj} and \cref{sec:boot}, we describe the particular instantiations of these two components in our demonstration of the framework. Finally, in \cref{sec:curriculum}, we show how to construct a self-supervised consistency loss to guide mask generation.

\subsection{CYBORGS Framework Abstraction}
\label{sec:overall}

Following typical contrastive learning frameworks in vision, we begin with a given RGB image $\mat{I} \in \mathbb{R}^{3 \times H \times W}$ of height $H$ and width $W$, and two transformations $t, t'$ independently sampled from data augmentation pipelines $\mathcal{T}, \mathcal{T'}$. For the time being, we assume we also have ground truth semantic segmentation masks $\{\mat{M}\} \in [0, 1]^{C \times H \times W}$. Each $H \times W$ binary mask $\mat{M}$ describes pixel-wise class membership for a particular class, for $C$ total classes. Applying the transformations to $\mat{I}, \{\mat{M}\}$ yields two augmented views $\vect{v} = t(\mat{I}), \vect{v'} = t'(\mat{I})$, and two semantic maps $\{\vect{m}\} = t(\{\mat{M}\}), \{\vect{m'}\} = t'(\{\mat{M}\})$. Note that every $\vect{m}$ contains object-level assignments spatially aligned with view $\vect{v}$, and likewise every $\vect{m'}$ aligns with $\vect{v'}$. After passing view $\vect{v}$ to a (fully) convolutional encoder $f_\theta$ for featurization, we can extract a (sub)set of intermediate feature maps $\{\mat{F}\} = \{\vect{y}^{[1]}, \dots, \vect{y}^{[l]}\}$, where $\vect{y}^{[l]} = f_\theta^{[l]}(\vect{v})$ for layer $l$. Doing the same for view $\vect{v'}$ yields feature maps $\{\mat{F'}\}$.

These feature maps inherently contain spatial and latent information about the image, which we can leverage using the segmentation masks. The core idea is conceptually simple and lightweight: we can sample arbitrary regions in the feature maps and apply the binary masks $\{\mat{m}\}, \{\mat{m'}\}$ to filter out groups of features which correspond to the same underlying object regions. Applying mean pooling, concatenation, or some other general aggregation operator to these groups yields feature vectors containing similar and dissimilar object-level semantics. These positive-negative pairs allow us to use a flexible class of contrastive objectives to train our encoder $f_\theta$. Note that this naturally requires upsampling or downsampling either the masks or the feature maps to the same spatiality, and our framework is entirely agnostic to these details. But a more immediate problem is obtaining reasonable masks $\{\mat{M}\}$ to begin with.

\begin{figure}[t]
\centering
\scalebox{0.8}{
\includegraphics[width=\textwidth]{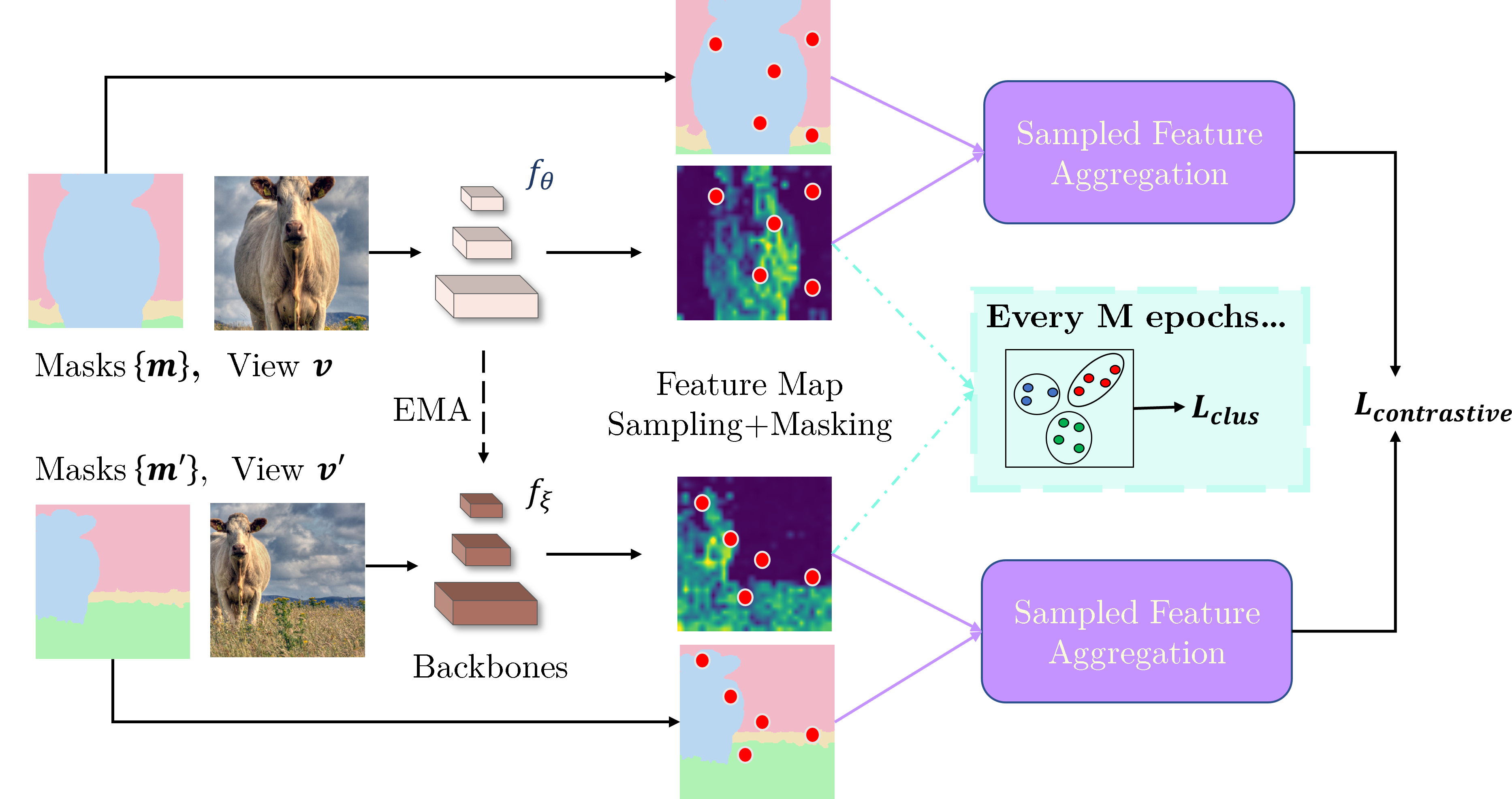}
}
\caption{\textbf{CYBORGS Training Framework.}
We sample over the feature maps for different views, using segmentation masks to identify similar and dissimilar object regions. These are aggregated into positive and negative feature vectors, respectively, for the contrastive objective $\mathcal{L}_{mask}$ (\cref{sec:mask_obj}). Periodically, we also backprop through a clustering consistency loss $\mathcal{L}_{clus}$ (\cref{sec:curriculum}).}
\label{fig:overall_fig}
\vspace{-15pt}
\end{figure}

A crucial assumption we have thus maintained is that ground truth segmentation masks are available. Indeed, without specification of how object regions correspond to each other across views, the very notion of positives and negatives for a contrastive formulation becomes ill-defined. Previous works which have relied on such masks in a similar fashion have used simple spatial heuristics such as grid-based masks, or more complex unsupervised algorithms such as graph cut segmentations \cite{bai2022point, zhang2020hierarchical, henaff2021efficient}. Ultimately, we find that these approaches yield unsatisfactory masks which are semantics-unaware, or require significant hand-tuning, especially when employed on scene images. But composing a learning-based procedure is non-trivial; the contrastive objective cannot backpropagate through the non-differentiable augmentations $t, t'$ and modify a mask $\vect{m}$ directly.

To this end, our framework bootstraps segmentation masks using representations from the partially trained model $f_\theta$. This idea is motivated by two insights. Firstly, the contrastive objective directly improves the encoder $f_\theta$, and thus leveraging the features from $f_\theta$ can help us obtain semantic-aware masks which correspondingly improve over the course of training. Secondly, recall that the ultimate goal of our framework is to improve representation learning. Since downstream transfer of representations takes places on $f_\theta$, using the representations from $f_\theta$ to construct our segmentation masks ensures that representation quality and bootstrapped mask quality are tied together. Implementation-wise, our framework is agnostic to the actual algorithm employed for mask generation, with the only constraint being that the method cannot rely on ground truth supervision. For concreteness, we illustrate in \cref{sec:boot} how to generate robust masks using a simple KMeans clustering-based algorithm on the feature maps $\{\mat{F}\}$. By iterating between contrastive updating of $f_\theta$ and unsupervised generation of masks, we mutally improve our representations and segmentations.

\subsection{Mask-Dependent Contrastive Learning}
\label{sec:mask_obj}

To demonstrate the utility of our framework, we first choose the loss function from \cite{henaff2021efficient} as the particular instantiation of a mask-based contrastive objective for training our encoder in the first stage. We provide a high level review here.


In \cite{henaff2021efficient}, $\{\mat{F}\}$ is a single $2048 \times 7 \times 7$ feature map extracted from the final layer of a standard ResNet-50 encoder processing view $\vect v$ (before average pooling). The entire feature map is sampled, and the segmentation masks in $\{\vect m\}$ are spatially downsampled accordingly. Aggregation of $\{\mat{F}\}$ is obtained via mask-based pooling for each $\vect m \in \{\vect m\}$:

\begin{equation}
\label{eqn:ft_pooling}
    \vect{h_m} = \frac{1}{\sum_{i, j} \vect{m}[i, j]} \sum \limits_{i, j} \vect{m}[i, j] \mat{F}[i, j]
\end{equation}

Feature map $\{\mat{F'}\}$ is similarly aggregated after processing view $\vect{v'}$ with a target encoder $f_\xi$, yielding $\vect{h'_{m'}}$. For additional asymmetry, $\vect{h_m}$ is further transformed by an online projector $g_\theta$ and predictor $q_\theta$ to obtain $\vect{v_m} = q_\theta(g_\theta(\vect{h_m}))$, and $\vect{h_{m'}}$ is transformed by a target projector $g_\xi$ to obtain $\vect{v'_{m'}} = g_\xi (h'_{m'})$. The target parameters $\xi$ are updated as an exponential moving average (EMA) of their online counterparts $\theta$. The final mask-based contrastive objective is given by:

\begin{equation}
\label{eqn:contrast}
  \mathcal{L}_{contrastive}  = \mathbb{E}_{\vect{m}, \vect{m'} \sim \{\vect{m}\}, \{\vect{m'}\}} \left[-\log \frac{\exp(\vect{v_m} \cdot \vect{v'_{m'}})}{\exp(\vect{v_m} \cdot \vect{v'_{m'}}) + \sum_n \exp(\vect{v_m} \cdot \vect{v_{n}})} \right]
\end{equation}
for negative pooled features $\{\vect{v}_n\}$ sampled from different masks and images.

In addition, inspired by prior art demonstrating that different layers within a CNN encode information at different semantic resolutions \cite{kornblith2019better, hariharan2015hypercolumns, Xu_2021_ICCV}, we also extract and utilize features from throughout ResNet-50, instead of relying solely on features from the final convolutional map as in \cite{henaff2021efficient}. By fusing these features together spatially (after upsampling or downsampling), downstream learning is able to leverage information across the semantic spectrum, from low-level local structure, to high-level global style. Further details are available in the appendix.

\subsection{Bootstrapping Segmentation Masks}
\label{sec:boot}
Recall that our framework is agnostic to the particular algorithm used in the second stage bootstrapping of better segmentation masks. For simplicity, we illustrate the details of this stage using a classic KMeans clustering algorithm.

\begin{wrapfigure}{R}{0.5\textwidth}
    \scalebox{0.5}{
        \includegraphics[width=\textwidth]{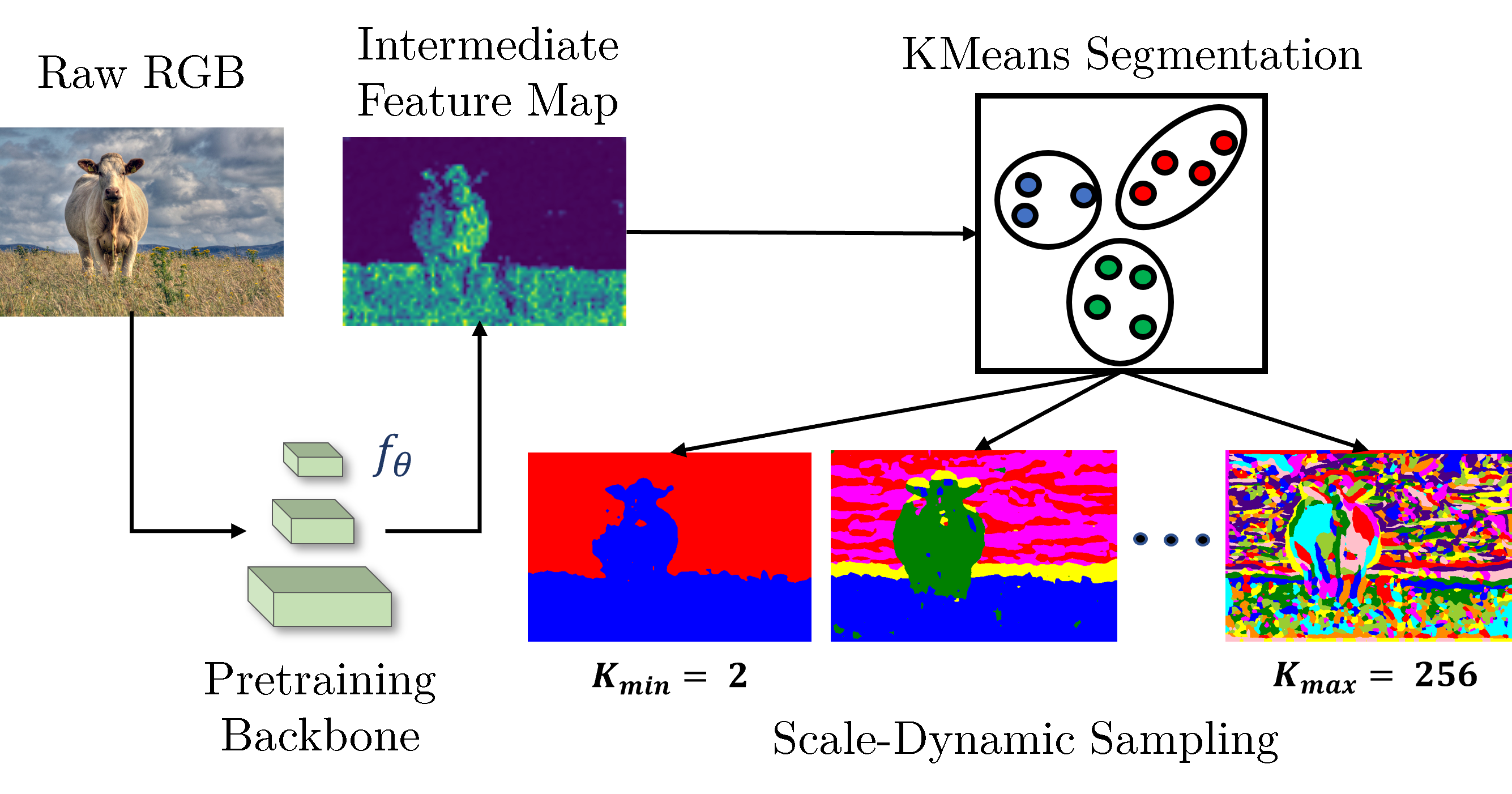}
    }
  \caption{\textbf{Bootstrapping Masks.} To generate the segmentation masks, we perform simple KMeans clustering on a feature map from the trained backbone, with a dynamic number of clusters.}
\label{fig:boot_fig}
\vspace{-15pt}
\end{wrapfigure}

More formally, we begin by considering a batch of $B$ input RGB images $\{\mat{I}\} \in \mathbb{R}^{B \times 3 \times H \times W}$, and a (fully) convolutional backbone $f_\theta$ which has been trained in a self-supervised fashion via the objective in \cref{eqn:contrast}. We choose a particular layer $\boldsymbol{\ell}$, and extract the feature map $\vect{y}^{[\boldsymbol{\ell}]}_\theta = f_\theta^{[\boldsymbol{\ell}]}(\{\mat{I}\}) \in \mathbb{R}^{B \times D_F \times H_F \times W_F}$. We omit the layer index $\boldsymbol{\ell}$ and online encoder parameters $\theta$ for brevity, so that $\vect{y} \triangleq \vect{y}^{[\boldsymbol{\ell}]}_\theta$. We then flatten the feature maps and $\ell2$-normalize feature-wise, generating a matrix of features $\mat{F} \in \mathbb{R}^{(B \cdot H_F \cdot W_F) \times D_F}$. Given a hyperparameter $K$, representing the number of clusters (or unique object classes) within the segmentation mask, we perform spherical $K$-means clustering on $\mat{F}$, ending up with a matrix of feature prototypes $\mat{P} = \{\vect{\mu}_1, \vect{\mu}_2, \dots, \vect{\mu}_K\} \in \mathbb{R}^{D_F \times K}$. We assign to each cell in the original feature map $\vect{y}^{[\boldsymbol{\ell}]}$ a cluster label based on their Euclidean distances to the prototypes in $\mat{P}$. Finally, we broadcast the class assignments back to the original dimensions of the image $\mat{I}$ via nearest neighbor interpolation, akin to \cite{chen2021intriguing}.

\paragraph{Periodic bootstrapping.}
Performing such a clustering operation on every epoch to regenerate the segmentations can be expensive. Even if computation was not an issue, we empirically find that representations do not improve monotonically with epochs, so bootstrapping masks too frequently can actually lead to worse masks. Moreover, as a result of an undertrained encoder $f_\theta$ at the beginning of training, we obtain poorer early clusterings; noisy masks lead to noisy gradients for updating the encoder, and vice versa. Thus, to avoid representation collapse, we periodically bootstrap the segmentation masks every $N$ epochs, where $N$ is a hyperparameter much greater than 1.

\paragraph{Scale-dynamic sampling.}

The choice of $K$ also merits discussion. Given access to some oracle, a natural choice might be to set $K$ equal to the number of unique object classes within the image. However, as a number of prior works have identified, the semantic context provided by extra ``distractor'' classes outside of the main object classes can serve as a useful signal for clustering \cite{ji2019invariant, caron2018deep, chen2021intriguing}. But increasing $K$ also requires more images within the bootstrapping batch to perform KMeans reliably on the features, reducing the scalability of our method.

To balance these motivations, for every batch of images where we wish to bootstrap segmentations, we dynamically sample integer $K$ uniformly between $K_{min} = 2$ and $K_{max} = 256$, inclusive. Intuitively, $K_{min} = 2$ represents a mask which imparts the model with simple foreground-background semantics, while the upper bound of $K_{max} = 256$ yields an oversegmentation (COCO offers only 81 labeled object segmentation classes.) By varying $K$ in such a fashion, not only do we maintain efficiency in bootstrapping, but we also reintroduce our model to information of varying semantic scale on every bootstrap cycle. As we show in \cref{sec:ablations}, this technique improves the robustness of our representations.

\subsection{Consistency as a Curriculum for Segmentation}
\label{sec:curriculum}

Despite the use of periodic bootstrapping and scale-dynamic sampling, we find that the long training schedules employed in contrastive learning can still lead to divergence between our representation learning and semantic segmentation objectives. This is because our framework up to now improves the segmentation only \emph{implicitly}. While we are optimizing on every iteration our contrastive objective in \cref{eqn:contrast}, regularly improving our encoded representations, the bootstrapping of masks is optimization-free with respect to the encoder. Without an update signal to explicitly encode the semantics of desirable vs. non-desirable segmentations, the encoder over-prioritizes the goal of representation learning, and can diverge from a feature distribution which yields good segmentations.

\paragraph{Clustering consistency.}
To this end, we reuse a universal paradigm in contrastive learning: similar objects across different scenes and different views should have similar labels. We introduce a clustering consistency loss, similar to that employed in \cite{cho2021picie}, which can be applied more regularly every $M$ epochs, where $M$ is more frequent than the every $N$ epochs used per bootstrapping cycle.

Concretely, recall the feature map $\vect y \in \mathbb{R}^{D_F \times H_F \times W_F}$ and feature prototypes $\mat{P} = \{\vect{\mu}_1, \vect{\mu}_2, \dots, \vect{\mu}_K\} \in \mathbb{R}^{D_F \times K}$ we obtained in \cref{sec:boot} after processing view $\vect v$ using the online encoder $f_\theta$. We obtain a similar map $\vect y'$ and set of prototypes $\mat{P'} = \{\vect{\mu'}_1, \vect{\mu'}_2, \dots, \vect{\mu'}_K\}$ after featurizing $\vect v'$ with the target encoder $f_\xi$. Consider the feature at pixel $[i, j]$ within $\vect{y}$, for an arbitrary $1 \leq i \leq H_F$ and $1 \leq j \leq W_F$. With a slight abuse of notation, we let $\vect{\mu}_{[i, j]}$ represent the prototype this feature is assigned to under $\mat{P}$ (and similarly, $\vect{\mu'}_{[i, j]}$ the assignment of $\vect{y'}[i, j]$ under $\mat{P'}$). Then we define a clustering consistency loss via:

\begin{equation}
    \label{eqn:consistency}
    \scalebox{0.9}{
    $\mathcal{L}_{clus} = \frac{1}{H_FW_F} \sum \limits_{i = 1}^{H_F} \sum \limits_{j = 1}^{W_F} \overbrace{d \left(\vect{\mu}_{[i, j]}, \vect{y}_{[i, j]} \right) + d \left(\vect{\mu'}_{[i, j]}, \vect{y'}_{[i, j]} \right)}^{intra-loss} +
    \overbrace{d \left(\vect{\mu'}_{[i, j]}, \vect{y}_{[i, j]} \right) + d \left(\vect{\mu}_{[i, j]}, \vect{y'}_{[i, j]} \right)}^{inter-loss}$
    }
\end{equation}

where $d(\cdot, \cdot)$ is some distance function. Intuitively, intra-cluster consistency enforces that under one scene, object regions with similar features should be clustered into similar prototypes. Similarly, inter-cluster consistency enforces that under different scenes, we still wish for features from different regions corresponding to similar objects to be assigned to the same prototype. This forces our learned prototypes to be invariant to differences between views and generalize to object-centric semantics, which translates readily to higher fidelity segmentation masks during bootstrapping updates.

To formulate $d(\cdot, \cdot)$, we draw inspiration from recent work which demonstrates that the infoNCE objective in contrastive learning promotes a feature space which is uniformly distributed on the unit hypersphere \cite{wang2020understanding}. The von Mises-Fisher (vMF) distribution defines a probability density over a unit hypersphere, making it a natural candidate to characterize the feature space learned by our mask-based contrastive objective in \cref{eqn:contrast}. We refer readers to a comprehensive treatment in \cite{gopal2014mises} for details. In our setting, we can assume a vMF mixture model where each feature $\vect{y}$ is drawn uniformly from one of $K$ vMF distributions, each parameterized by a feature clustering prototype $\vect{\mu}_1, \vect{\mu}_2, \dots, \vect{\mu}_K$, and sharing a common concentration hyperparameter $\kappa$. Then our clustering consistency loss objective is formulated as maximizing the posterior likelihood of a particular encoded feature $\vect{y}$ being assigned to its corresponding cluster $c$ under this mixture, with $1 \leq c \leq K$. That is, we seek to minimize the negative log-likelihood given by:

\begin{equation}
\label{eqn:vmf_clustering}
d(\vect{\mu}_{[i, j]}, \vect{y}) = - \log p(\vect{\mu}_{[i, j] = c} \mid \vect{y}, \vect{\mu}_1, \vect{\mu}_2, \dots, \vect{\mu}_K) = -\log \frac{\exp \left(\kappa \vect{\mu}_{[i, j]}^T \vect{y} \right)}{\sum \limits_{c' = 1}^{K} \exp \left(\kappa \vect{\mu}_{c'}^T \vect{y}\right)}
\end{equation}

The vMF clustering loss objective described in \cref{eqn:consistency} also serves an additional purpose towards the beginning of our pretraining pipeline. In the total absence of reliable masks before the first bootstrapping cycle, we train our encoder $f_\theta$ strictly with the loss in \cref{eqn:consistency}, setting $K$ to a fixed parameter depending on the median number of objects per scene in our dataset (\eg for COCO, we use $K = 8$). This \emph{vMF warmup period} of $W$ epochs ($W = 5$ in our work) serves to burn in our encoder. After more reasonable representations have been learned, we immediately bootstrap the masks, and subsequent epochs using a combination of the mask-based contrastive loss in \cref{eqn:contrast} and the vMF clustering loss in \cref{eqn:consistency}, as their respective periods $N$ and $M$ dictate. For a comprehensive outlining of our algorithm flow, we refer readers to the pseudocode presented in the appendix.

\section{Experiments}

In our experiments, we aim to demonstrate that joint learning of general representations and semantic segmentation can be successfully accomplished via our bootstrapping method. We show strong performance on multiple downstream tasks (\cref{sec:main_results}), surprisingly robust segmentation performance over a long-tailed distribution of objects (\cref{sec:seg_results}), and a convincing array of ablations which validate our design choices and methodological contributions (\cref{sec:ablations}).

\subsection{Experimental Settings}
\label{sec:exp_settings}
\paragraph{Datasets.}

Given our primary goal of learning on images in the wild, we follow previous works \cite{selvaraju2021casting, wang2021dense, liu2020self} and pretrain on the train2017 split of the MS COCO dataset \cite{lin2014microsoft}. With $\sim$118k images of natural settings, MS COCO is widely adopted as a benchmark more reflective of real-world scenarios across a breadth of downstream tasks of interest, such as object detection or instance segmentation. For a relevant quantitative comparison, note that the heavily object-dominant ImageNet dataset contains on average 1.1 objects per image, whereas the average scene image in COCO contains 7.3 objects \cite{wang2021dense}. Crucially, we use no scene-level, object-level, or pixel-level label information in our pretraining pipeline.

\paragraph{Implementation details.}
To enable easy comparison to other SSL works in similar settings \cite{selvaraju2021casting, xie2021unsupervised, liu2020self, wang2021dense}, we use a ResNet-50 backbone in all of our models. Other architectural details such as the dimensionality of projection and prediction MLPs described in \cref{sec:mask_obj} follow directly from BYOL \cite{grill2020bootstrap}.

For our mask-based contrastive objective in \cref{eqn:contrast}, we aggregate features from $\texttt{res2}, \texttt{res3}, \texttt{res4}$, downsampling all layers to a spatial resolution of $7 \times 7$. This allows us to leverage a lightweight but comprehensive semantic hierarchy. We bootstrap the segmentation masks every $N = 100$ epochs, performing clustering on the feature map from $\texttt{res2.b2}$ in batches of 16 images (where $\texttt{b2}$ refers to block 2). We use a vMF warmup period of 5 epochs; outside the warmup period, the vMF clustering loss is employed every 5 epochs with weight $\lambda = 0.1$ and $\kappa = 10$. In pretraining, we use the LARS optimizer \cite{you2019large} with a batch size of 64 across 8 NVIDIA RTX 3090s for 800 epochs. The initial learning rate is set to 0.1, and the weight decay is $1.5e^{-6}$. Clustering is implemented via GPU-accelerated mini-batch approximation using the FAISS library \cite{johnson2019billion}.

\subsection{Main Results: Representation Learning}
\label{sec:main_results}
\begin{table*}[t]
    \newcommand{\apbbox}[1]{AP$^\text{bb}_\text{#1}$}
    \newcommand{\apmask}[1]{AP$^\text{mk}_\text{#1}$}
    \newcolumntype{Y}{>{\raggedright\arraybackslash}X}
    \newcolumntype{Z}{>{\centering\arraybackslash}X}

    \footnotesize
    \setlength\tabcolsep{1pt}
    \renewcommand{\arraystretch}{1.2}

    \scalebox{0.73}{
        \begin{tabular*}{\linewidth}{c l c c c cc c cc c ccc c cccccc}
        \addlinespace[-\aboverulesep] 
        \cmidrule[\heavyrulewidth]{1-21}
        ~
        & \multicolumn{1}{l}{\bf \multirow[b]{2}{*}{Method}}
        & ~
        & \multicolumn{1}{c}{\bf VOC07 clf.}
        &~& \multicolumn{2}{c}{\makecell{\bf IN-1k,\\ \bf 1\% Labels}}
        &~& \multicolumn{2}{c}{\makecell{\bf IN-1k,\\ \bf 10\% Labels}}
        &~& \multicolumn{3}{c}{\makecell{\bf VOC Detection \\}}
        &~& \multicolumn{6}{c}{\makecell{\bf COCO Instance Segmentation}} \\
        \cmidrule{4-4} \cmidrule{6-7} \cmidrule{9-10} \cmidrule{12-14} \cmidrule{16-21}
    
        ~ & ~ & ~ & \multicolumn{1}{c}{mAP} & ~ & \multicolumn{1}{c}{\makecell{Top-1\\ acc.}} & \multicolumn{1}{c}{\makecell{Top-5\\ acc.}} & ~ &
        \multicolumn{1}{c}{\makecell{Top-1\\ acc.}} & \multicolumn{1}{c}{\makecell{Top-5\\ acc.}} & ~ &
                    \apbbox{} & \apbbox{50} & \apbbox{75} &&
                    \apbbox{} & \apbbox{50} & \apbbox{75} &
                    \apmask{} & \apmask{50} & \apmask{75} \\
        \addlinespace[-\aboverulesep] 
        \cmidrule[\lightrulewidth]{1-21}
        \ttbf{1)}  & \kw{SimCLR} \cite{chen2020simple} &&
                    78.1 && 
                    23.4 & 46.4 &&
                    52.2 & 77.4 &&
                    -- & -- & -- &&
                    37.0 & 56.8 & 40.3 & 33.7 & 53.8 & 36.1 \\
    
        \ttbf{2)}  & \kw{MoCo-v2} \cite{chen2020improved} &&
                    82.2 && 
                    28.2 & 54.7 && 
                    57.1 & 81.7 &&                            
                    54.7 & 81.0 & 60.6 &&                       
                    38.5 & 58.1 & 42.1 & 34.8 & 55.3  & 37.3 \\ 
        
        \ttbf{3)}  & \kw{BYOL} \cite{grill2020bootstrap} &&
                    84.5 && 
                    28.4 & 55.9 && 
                    58.4 & 82.7 &&                            
                    55.5 & 81.7 & 61.7 &&                       
                    39.5 & 59.3 & 43.2 & 35.6 & 56.5 & 38.2 \\ 
        
        \ttbf{4)} & \kw{Bai et al.} \cite{bai2022point} &&
                    -- && 
                    -- & -- &&
                    -- & -- &&
                    57.1 & 82.1 & 63.8 && 39.8 & 59.6 & 43.7 & 35.9 & 56.9 & 38.6 \\ 
        
        \ttbf{5)}  & \kw{DenseCL} \cite{wang2021dense} &&
                    83.8 && 
                    -- & -- && 
                    -- & -- &&                            
                    56.7 & 81.7 & 63.0 &&                       
                    39.6 & 59.3 & 43.3 & 35.7 & 56.5 & 38.4 \\ 
        
        \ttbf{6)}  & \kw{CAST} \cite{selvaraju2021casting} &&
                    73.1 && 
                    -- & -- && 
                    -- & -- &&                            
                    54.2 & 80.1 & 59.9 &&                       
                    36.7 & 56.7 & 39.9 & 33.6 & 53.6 & 35.8 \\ 
        
         \ttbf{7)} & \kw{ORL} \cite{xie2021unsupervised} &&
                    86.7 && 
                    31.0 & 58.9 && 
                    60.5 & \bf 84.2 &&                            
                    55.8 & 82.1 & 62.3 &&                       
                    40.3 & 60.2 & 44.4 & 36.3 & 57.3 & 38.9 \\ 
        
        \addlinespace[-\aboverulesep] 
        \cmidrule[\lightrulewidth]{1-21}
         \ttbf{8)}  & \app (ours) &&
                    \bf 86.9 && 
                    \bf 31.3 & \bf 59.4 &&
                    \bf 61.7 & \bf 84.2 &&                            
                    \bf 58.0 & \bf 83.0 & \bf 64.3 &&                       
                    \bf 42.0 & \bf 62.6 & \bf 46.2 & \bf 38.0 & \bf 59.7 & \bf 40.8 \\
        
        \addlinespace[-\aboverulesep] 
        \cmidrule[\heavyrulewidth]{1-21}
        \end{tabular*}
        }
    \setlength{\abovecaptionskip}{5pt plus 3pt minus 2pt}
    \caption{\textbf{Transfer Learning on Downstream Tasks.} We report strong, state-of-the-art performance across linear classification on VOC07, semi-supervised finetuning on ImageNet-1k, and transfer on VOC object detection and COCO instance segmentation. All methods are pretrained on COCO with a ResNet-50 backbone, and finetuned on the reported datasets.}
    \label{tab:main_results}
    \vspace{-15pt}
\end{table*}

We follow standard downstream transfer-based protocols to evaluate the strength of representations learned by \app. In particular, we begin with \emph{frozen} linear evaluation on VOC07 and semi-supervised transfer on ImageNet-1k. In comparison to similar state-of-the-art self-supervised methods pretrained on COCO, we achieve improvements of $+0.2$ mAP for VOC07 and $+0.3\%, +1.2\%$ in top-1 accuracy for semi-supervised 1\% and 10\% on IN-1k, respectively. While these gains are only incremental, image classification requires semantic-level knowledge~\cite{purushwalkam2020demystifying, yang2021instance, ballard2021hierarchical}, whereas we design our method around leveraging pixel-level information, and so even marginal gains are a surprising windfall.

While linear probing has been treated as the gold standard for assessing feature quality, that strong performance in tasks such as detection and segmentation are even more reflective of potent learned representations. We demonstrate convincing state-of-the-art on PASCAL VOC detection and COCO instance segmentation. In comparison to a strong and well-established BYOL baseline, we provide a +2.5 AP improvement on the former, and a +2.5 and +2.4 AP improvement on the latter. Our results on segmentation in particular are noteworthy; while we do not make use of pixel annotations, our bootstrapping scheme clearly aligns with latent information critical to the segmentation task. To further verify this robust segmentation performance, we also perform transfer-based evaluation on CityScapes semantic segmentation \cite{cordts2016cityscapes}, as well as LVIS long-tailed instance segmentation \cite{gupta2019lvis}, which we detail in the appendix. We achieve state-of-the-art on these datasets amongst all other previous SSL methods pretrained on COCO, with a substantial +3.4 AP and +3.3 AP improvement on LVIS, demonstrating that our framework can also generalize to unseen object structures and semantics.

\subsection{Segmentation Quality}
\label{sec:seg_results}
\begin{figure}[t]
\centering
\begin{tabular}{cccc}
\subfloat{\includegraphics[width=0.24\textwidth]{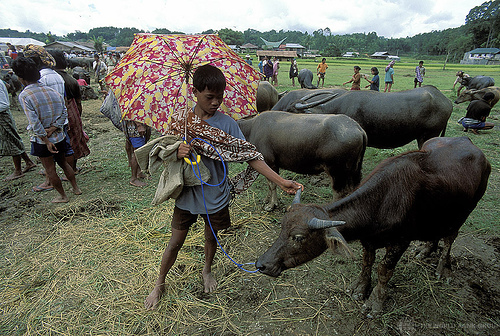}} &
\subfloat{\includegraphics[width=0.24\textwidth]{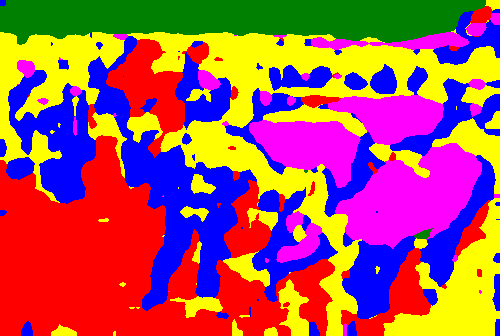}} &
\subfloat{\includegraphics[width=0.24\textwidth]{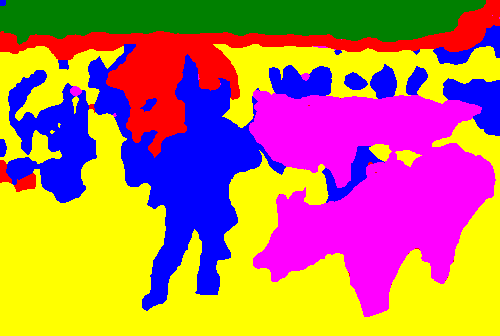}} &
\subfloat{\includegraphics[width=0.24\textwidth]{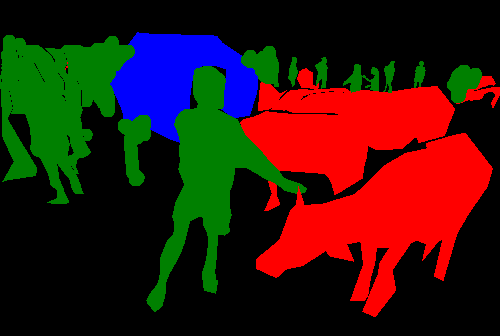}} \\
\\[-16pt] & \\[-16pt] & \\[-16pt] & \\[-16pt] \\
\subfloat{\includegraphics[width=0.24\textwidth]{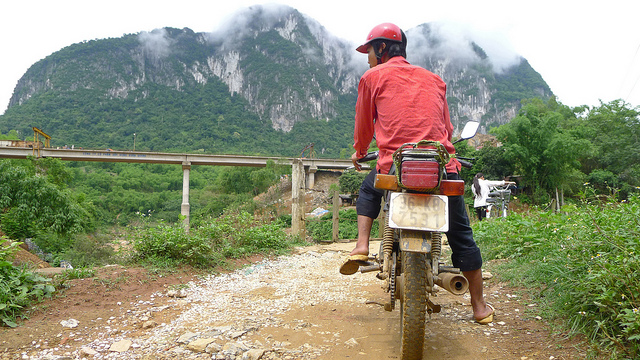}} &
\subfloat{\includegraphics[width=0.24\textwidth]{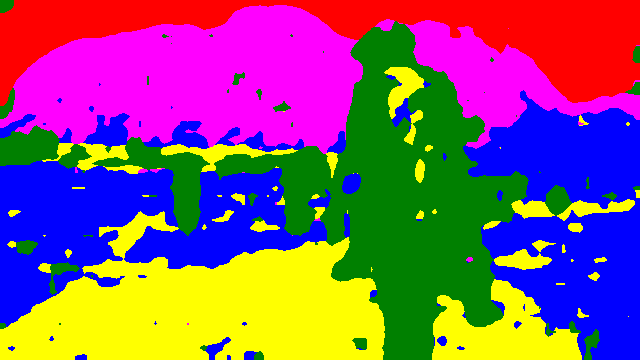}} &
\subfloat{\includegraphics[width=0.24\textwidth]{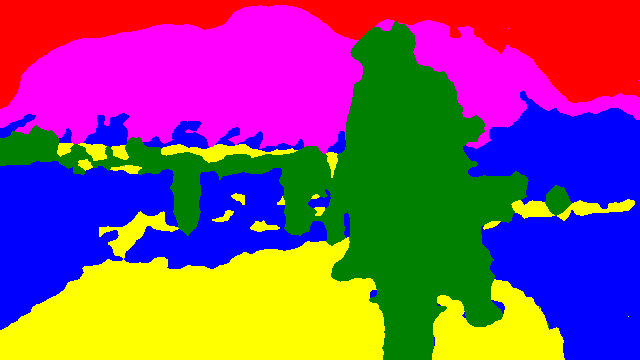}} &
\subfloat{\includegraphics[width=0.24\textwidth]{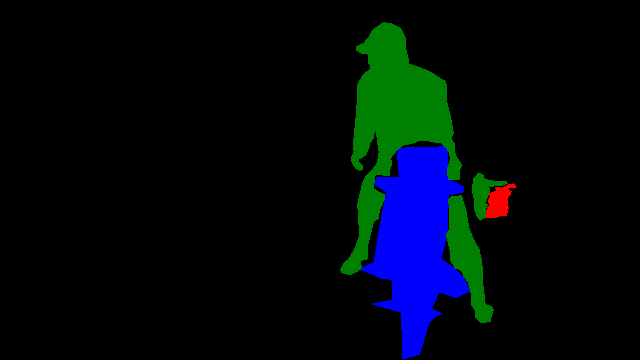}} \\
\\[-16pt] & \\[-16pt] & \\[-16pt] & \\[-16pt] \setcounter{subfigure}{0} \\
\subfloat[Raw RGB]{\includegraphics[width=0.24\textwidth]{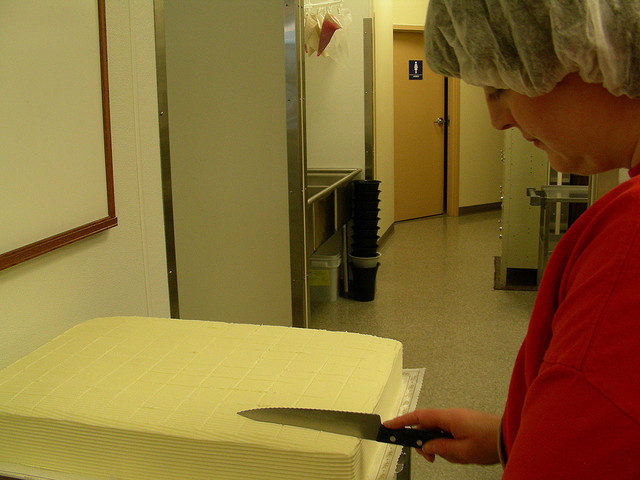}} &
\subfloat[KMeans Mask]{\includegraphics[width=0.24\textwidth]{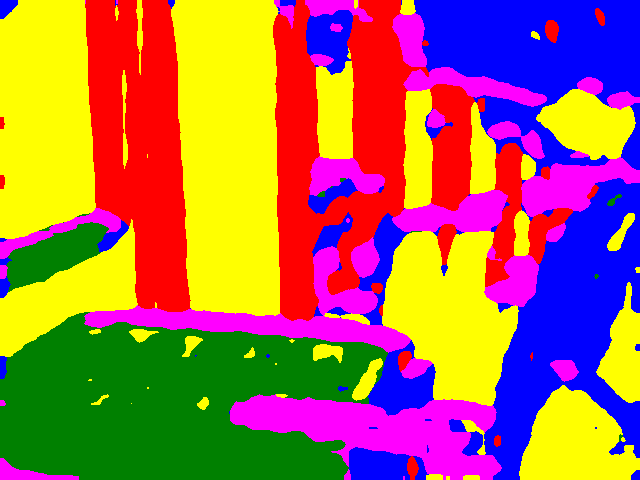}} &
\subfloat[CRF-Refined Mask]{\includegraphics[width=0.24\textwidth]{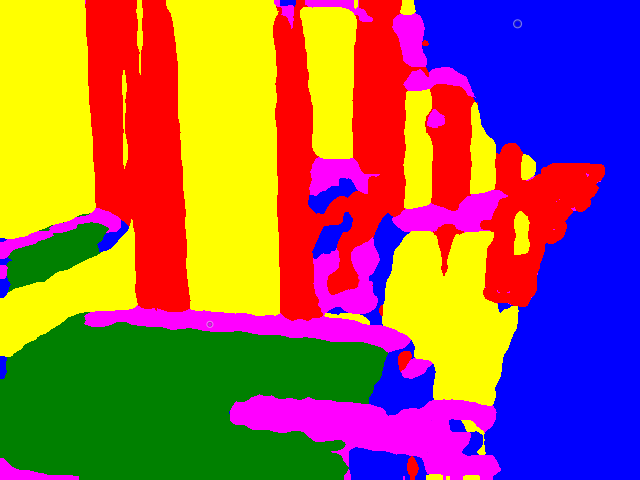}} &
\subfloat[Ground Truth]{\includegraphics[width=0.24\textwidth]{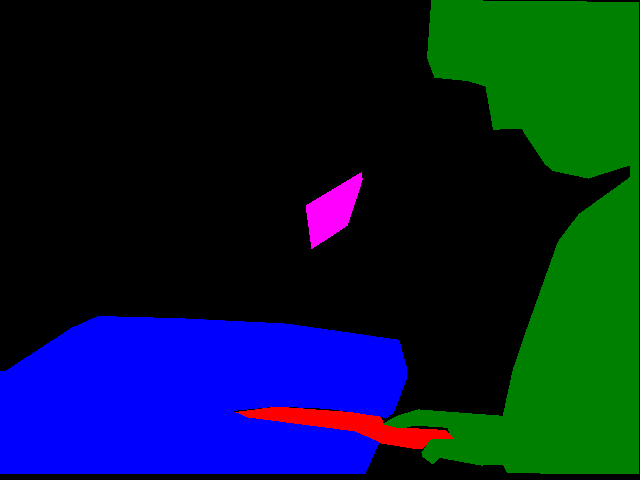}}
\end{tabular}
\caption{\textbf{Bootstrapped segmentation masks from a CYBORGS-pretrained encoder on COCO.} We show KMeans segmentations on the bilinearly upsampled feature maps for visual quality. During actual bootstrapping, we first segment the feature map, before performing nearest neighbors upsampling, and do not perform CRF refinement on the mask. Colors do not necessarily correspond across images (rows) or between mask types (columns), but are consistent within a single image itself. }
\label{fig:seg_results}
\end{figure}

We first confirm qualitatively (\cref{fig:seg_results}) that bootstrapped masks generated by \app{} are indeed semantically meaningful. Note that our clustering-based segmentations easily extend beyond the original labeled classes of COCO, despite receiving no ground truth information about pixel labels throughout pretraining.

\paragraph{How does \app{} work with such noisy masks?} In addition to the masks generated by clustering on the feature maps from the backbone encoder, we also show the mask resulting from refinement using a fully connected conditional random field (CRF), using the distances to feature prototypes in latent space as priors, following the protocol described in previous works \cite{krahenbuhl2011efficient, chen2017rethinking}. We argue that although the raw masks at a pixel-level appear to be noisy, their easy refinement into masks closely aligned with ground truth masks indicates that the encoded features are quite well aligned with object-level concepts at the semantic level.

\begin{figure}[t]
    \centering
    \scalebox{0.8}{
        \captionsetup[subfloat]{captionskip=16pt}
        \subfloat[\bf{Importance of bootstrapping.}]{
            \adjustbox{valign=c}{
            \footnotesize
            \begin{tabular}{lr}
                \toprule
                    \textbf{Type of Mask} & \textbf{mIoU} \\ 
                \midrule
                Random Crop & 15.9 \\ 
                5x5 Grid & 18.7 \\ 
                FH Masks & 27.7 \\
                Obj. Bounding Boxes & 29.0 \\ 
                \app{} Masks (ours) & 33.6 \\ 
                Ground Truth Masks & 35.2 \\ 
                \bottomrule
            \end{tabular}
            \label{tab:mask_type}
            }
        }
        \captionsetup[subfloat]{captionskip=1pt}
        \subfloat[\textbf{Masks (\textcolor{NavyBlue}{blue}) and representations (\textcolor{red}{red}) improve jointly.}]{
            \includegraphics[width=0.45\textwidth,valign=c]{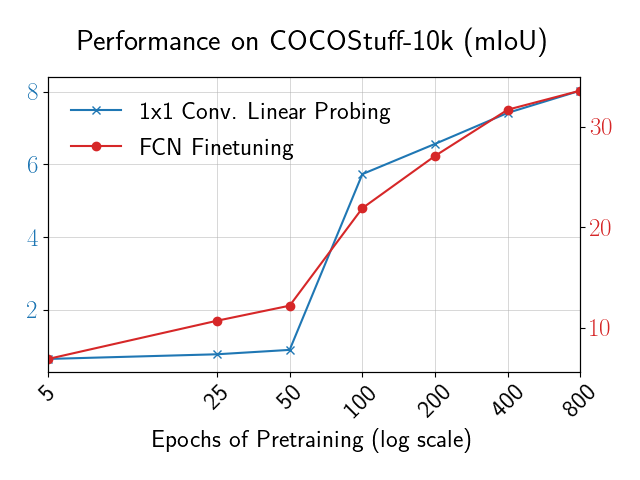} 
            \label{fig:boot_improve}          
        }
    }
    \caption{\textbf{Using semantic segmentation performance on COCO-Stuff-10k to evaluate bootstrapping value.} (a) Replacing our bootstrapping segmentation core with static boxes from other unsupervised heuristics leads to decreased performance. (b) Note a single epoch of using improved masks can lead to significant gains (epoch 100).
    }
    \vspace{-15pt}
\end{figure}

\paragraph{Why bootstrap masks?} To further demonstrate the robustness of our bootstrapping process for mask generation, we retrain \app{} using alternative masks. Instead of bootstrapping masks, we employ random cropping masks (i.e. all pixels in the scene belong to the same class), a $5 \times 5$ spatial grid mask and Felzenszwalb-Huttenlocher (FH) masks used in \cite{henaff2021efficient, bai2022point}, detection-level object masks acquired via selective search pre-processing, and ground truth masks available in COCO. These masks are generated before pretraining and remain fixed, supplanting our bootstrapping algorithm. Given the unsupervised generalization of masks generated under our framework to a long-tailed distribution of objects (c.f. \cref{fig:seg_results}), we evaluate the representations by transferring the trained backbones to a ResNet-50 FCN and finetuning end-to-end on COCO-Stuff-10k semantic segmentation. COCO-Stuff-10k is a \emph{densely} labeled subset of COCO, comprising of 9k images for training and 1k images for testing, across 171 semantic categories \cite{caesar2018coco}. We verify in \cref{tab:mask_type} that bootstrapping mask-level information through \app{} outperforms detection-level boxes obtained from selective search, and nears performance of pretraining with fixed, stable ground truth masks.

\paragraph{Joint improvement of masks and representations.} The harmonious interplay between the representation learning and semantic segmentation components of our framework is one of our major contributions. To ascertain that representations and segmentation quality mutually improve over pretraining, we continue to assess semantic segmentation performance on the COCO-Stuff-10k dataset, for saved checkpoints throughout various stages of pretraining. For a batch of input images, we extract frozen feature maps from the same layer we use to bootstrap segmentation masks (\texttt{res2.b2}), and bilinearly interpolate to the original image dimensions. We then add a single layer of 1x1 convolutions to predict the pixel labels, yielding a final setup akin to linear probing in transfer-based evaluation.

Because only this last layer is trainable in the resulting model, segmentation performance is heavily dependent on the quality of the extracted feature maps. Since these are exactly the inputs to our KMeans segmentation algorithm, we obtain transfer results which correlate readily with the quality of our bootstrapped masks. To evaluate representation quality in the same pretrained models, we transfer the ResNet-50 backbones, unfreeze all layers, and add an FCN head, finetuning on COCO-Stuff-10k end-to-end. We perform these evaluations for \app{} models pretrained for 5, 25, 50, 100, 200, 500 and 800 epochs. As seen in \cref{fig:boot_improve}, this evaluation scheme demonstrates that mask quality and representation quality improve jointly over the course of pretraining. Note that we bootstrap masks for the first time at the \emph{beginning} of epoch 100 using our partially pretrained backbone; a subsequent iteration over the entire dataset is sufficient to improve both mask and representation semantics significantly.

\subsection{Ablations and Discussion}
\label{sec:ablations}
All ablation models are pretrained using a ResNet-50 backbone, and evaluations are performed on PASCAL VOC detection for faster turnaround time.

\begin{table*}[t]
    \newlength\savewidth\newcommand\shline{\noalign{\global\savewidth\arrayrulewidth
      \global\arrayrulewidth 1pt}\hline\noalign{\global\arrayrulewidth\savewidth}}
    \newcommand{\tablestyle}[2]{\setlength{\tabcolsep}{#1}\renewcommand{\arraystretch}{#2}\centering\ftsize}
    \renewcommand{\paragraph}[1]{\vspace{1.25mm}\noindent\textbf{#1}}
    \newcommand\blfootnote[1]{\begingroup\renewcommand\thefootnote{}\footnote{#1}\addtocounter{footnote}{-1}\endgroup}
    \definecolor{baselinecolor}{gray}{.9}
    \newcommand{\baseline}[1]{\cellcolor{baselinecolor}{#1}}
    \newcommand{\apbbox}[1]{AP$_\text{\tiny #1}$}
    
    \newcolumntype{x}[1]{>{\centering\arraybackslash}p{#1pt}}
    \newcolumntype{y}[1]{>{\raggedright\arraybackslash}p{#1pt}}
    \newcolumntype{z}[1]{>{\raggedleft\arraybackslash}p{#1pt}}

\vspace{-.2em}
\centering
\ftsize
\subfloat[
\ftsize
\textbf{Bootstrapping frequency}. Mask bootstrapping too often or not enough leads to poor performance.
\label{tab:ablation_boot_freq}
]{
\centering
\ftsize
\begin{minipage}{0.19\linewidth}{\begin{center}
\begin{tabular}{y{18}x{18}}
N & \apbbox{all} \\
\shline
0 & 9.75 \\
10 & 52.6 \\
50 & 58.3 \\
100 & \baseline{58.0} \\
200 & 55.0 \\
400 & 52.7
\end{tabular}
\end{center}}\end{minipage}
}
\hspace{2em}
\subfloat[
\ftsize
\textbf{CRF in bootstrapping}. Refining bootstrapped masks with CRFs during pretraining is not necessary.
\label{tab:ablation_crf}
]{
\centering
\ftsize
\begin{minipage}{0.19\linewidth}{\begin{center}
\begin{tabular}{y{30}x{18}}
Case & \apbbox{all} \\
\shline
CRF & 58.6 \\
No CRF & \baseline 58.0 \\
~ & ~ \\
~ & ~ \\
~ & ~ \\
~ & ~
\end{tabular}
\end{center}}\end{minipage}
}
\hspace{2em}
\subfloat[
\ftsize
\textbf{Features in contrastive objective}. Leveraging a semantic hierarchy of features is important.
\label{tab:ablation_multi_contrast}
]{
\begin{minipage}{0.19\linewidth}{\begin{center}
\begin{tabular}{y{30}x{18}}
Layers & \apbbox{all} \\
\shline
2 & 55.1 \\
3 & 55.8 \\
4 & 52.7 \\
2+4 & 56.4 \\
2+3+4 & \baseline 58.0 \\
~ & ~
\end{tabular}
\end{center}}\end{minipage}
}
\hspace{2em}
\subfloat[
\ftsize
\textbf{Features in KMeans}.
Earlier maps are more amenable to KMeans segmentation.
\label{tab:ablation_multi_seg}
]{
\begin{minipage}{0.19\linewidth}{\begin{center}
\begin{tabular}{y{30}x{18}}
Layers & \apbbox{all} \\
\shline
2.b2 & \baseline 58.0  \\
2+3 & 58.2 \\
2+4 & 54.4 \\
2+3+4 & 55.0 \\
~ & ~ \\
~ & ~
\end{tabular}
\end{center}}\end{minipage}
}
\\
\centering
\vspace{.3em}
\subfloat[
\ftsize
\textbf{Scale-dynamic sampling}. Dynamically sampling cluster resolution for KMeans segmentation works best.
\label{tab:ablation_scale}
]{
\begin{minipage}{0.19\linewidth}{\begin{center}
\begin{tabular}{y{50}x{18}}
K & \apbbox{all} \\
\shline
$K = 2$ & 22.6 \\
$K = 81$ & 42.5 \\
$K = 256$ & 33.2 \\
$K \sim \mathcal{U}[2, 256]$ & \baseline 58.0 \\
~ & ~
\end{tabular}
\end{center}}\end{minipage}
}
\hspace{2em}
\subfloat[
\ftsize
\textbf{Clustering loss}. Euclidean distance for \cref{eqn:vmf_clustering} leads to significant performance degradation.
\label{tab:ablation_vmf_eucl}
]{
\begin{minipage}{0.19\linewidth}{\begin{center}
\begin{tabular}{y{36}x{18}}
Loss & \apbbox{all} \\
\shline
Euclidean & 53.8 \\
vMF & \baseline 58.0 \\
~ & ~ \\
~ & ~ \\
~ & ~
\end{tabular}
\end{center}}\end{minipage}
}
\hspace{2em}
\subfloat[
\ftsize
\textbf{vMF Loss Frequency}. Applying the vMF curriculum more regularly leads to stronger performance.
\label{tab:ablation_vmf_freq}
]{
\centering
\begin{minipage}{0.19\linewidth}{\begin{center}
\begin{tabular}{y{18}x{18}}
M & \apbbox{all} \\
\shline
0 & 41.8 \\
1 & 58.6 \\
5 & \baseline 58.0 \\
10 & 57.8 \\
50 & 55.6
\end{tabular}
\end{center}}\end{minipage}
}
\hspace{2em}
\subfloat[
\ftsize
\textbf{vMF Loss Weight}. Performance is sensitive to the presence but not weight of the vMF loss.
\label{tab:ablation_vmf_weight}
]{
\begin{minipage}{0.19\linewidth}{\begin{center}
\begin{tabular}{y{36}x{18}}
$\lambda$ & \apbbox{all} \\
\shline
0 & 41.5 \\
0.001 & 42.3 \\
0.01 & 57.8 \\
0.1 & \baseline 58.0 \\
1 & 57.2 \\
\end{tabular}
\end{center}}\end{minipage}
}
\setlength{\abovecaptionskip}{10pt plus 3pt minus 2pt}
\caption{\ftsize \textbf{Ablations} for design choices in \app{}. We report average precision (AP) for object detection on PASCAL VOC \texttt{test2007}. Default settings corresp. to \cref{tab:main_results} are highlighted in \colorbox{baselinecolor}{gray}.}
\label{tab:ablations}
\vspace{-15pt}
\end{table*}

\paragraph{Bootstrapping frequency.}

We perform a sensitivity analysis on the bootstrapping frequency parameter $N$, where we regenerate the masks on epoch $N, \,\, 2N, \dots$, using feature maps from the improving encoder (\cref{tab:ablation_boot_freq}). Using only the initial masks obtained under vMF warmup for pretraining (\ie $N = 0$) leads to collapsed performance. Moreover, bootstrapping the masks too frequently ($N$ = 10) also leads to a performance drop, consistent with our hypothesis in \cref{sec:boot} that unstable masks which are changing too rapidly can lead to representational collapse. Finally, we also note that bootstrapping too \emph{infrequently} (\ie $N = 400$) is similarly suboptimal, validating our default chosen schedule.

\paragraph{CRF-refinement of masks.}

Given the qualitative improvements of the CRF-refined masks when performing the final evaluation (c.f. \cref{fig:seg_results}), a natural consideration is to apply CRF post-processing to the masks during every bootstrapping cycle. As we show in \cref{tab:ablation_crf}, this brings only incremental improvements to the resulting representations, which we believe do not justify the increase in computational complexity. This result also further validates our claim in \cref{sec:seg_results} that the representations under \app{} are already well-aligned in latent space with respect to the semantic segmentation task.

\paragraph{Usage of multiple layers of features.}

Throughout our method, there are two points where we potentially use feature maps from multiple layers of our encoder backbone. The first is in the aggregation of features for our contrastive objective in \cref{eqn:contrast}. We show in \cref{tab:ablation_multi_contrast} that using $\texttt{res2}, \texttt{res3}, \texttt{res4}$ from our backbone in combination is crucial to performance. This further verifies that leveraging information from across the semantic spectrum learned by the encoder is vital.

The second point is in the bootstrapping of masks, where we use only the feature map from $\texttt{res2.b2}$ of our backbone. We show in \cref{tab:ablation_multi_seg} that feature aggregation across multiple layers does not help here. One explanation for this phenomenon is curse of dimensionality; a simple KMeans clustering procedure on extremely high dimensional features aggregated across multiple layers may result in clusters with few or no points.

\paragraph{Scale-dynamic sampling.}

We also perform an ablation on dynamically sampling the semantic resolution of masks during bootstrapping. We compare with a foreground-background masks ($K = 2$), object-level masks ($K = 81$ categories from COCO), and clustering with the same number of unique labels as the default graph-based segmentation algorithm used in \cite{henaff2021efficient}. As shown in \cref{tab:ablation_scale}, fixing the KMeans cluster dimension at any level reduces the performance of \app{}. This validates our choice to provide the encoder with diverse levels of detail through bootstrapped masks of varying semantic resolution.

\paragraph{vMF clustering loss.} We verify several properties of the clustering loss in \cref{eqn:vmf_clustering}. As seen in \cref{tab:ablation_vmf_eucl}, basing the loss on the vMF distribution, which aligns better with our hyperspherically-distributed embeddings, results in better transfer performance. We also examine the frequency at which the vMF clustering loss is applied, and how sensitive our method is to the weight of this loss. \cref{tab:ablation_vmf_freq} validates that applying the loss more frequently increases downstream transfer performance. In combination with \cref{tab:ablation_vmf_weight} we note the weight of the loss does not dramatically influence performance, but its presence is important; at $\lambda = 0, 0.001$ or if $M = 0$ (\ie no application of vMF loss), the performance collapses.

\section{Conclusion}

We have proposed \app{}, a novel self-supervised framework which learns object-level representations and semantic segmentation jointly, in an end-to-end fashion. In pretraining on complex scene images, our representations transfer competitively to a diverse array of downstream tasks, with particularly strong alignment with a long-tailed distribution of object-level segmentation semantics.

\noindent\textbf{Acknowledgements:} YG is supported by the Ministry of Science and Technology of the People's Republic of China, the 2030 Innovation Megaprojects ``Program on New Generation Artificial Intelligence'' (Grant No. 2021AAA0150000). YG is also supported by a grant from the Guoqiang Institute, Tsinghua University. RW would like to thank Yu Sun and Yingdong Hu for valuable edits to the paper, without which this work would not be possible.

\clearpage
\bibliographystyle{splncs04}
\bibliography{references}

\end{document}